\documentclass[11pt,a4paper]{article}
\usepackage[hyperref]{acl2020}
\usepackage{times}
\usepackage{latexsym}
\usepackage{graphicx}

\usepackage{tabu,multirow}
\usepackage{todonotes}
\usepackage{newunicodechar}
\newunicodechar{ℝ}{\mathbb{R}}

\usepackage{microtype}

\aclfinalcopy 

\title{$\mathbf{{R}^3}$: Reverse, Retrieve, and Rank for Sarcasm Generation\\ with Commonsense Knowledge}

\author{Tuhin Chakrabarty\textsuperscript{1,2}\thanks{~~The research was conducted when the author was at USC/ISI.}, 
  Debanjan Ghosh\textsuperscript{3}, 
  Smaranda Muresan\textsuperscript{2,4}
  \textbf{and} \textbf{Nanyun Peng}\textsuperscript{1}\\ 
  \textsuperscript{1}Information Sciences Institute, University of Southern California \\
  \textsuperscript{2}Department of Computer Science, Columbia University \\
  \textsuperscript{3}Educational Testing Service,
  \textsuperscript{4}Data Science Institute, Columbia University\\\AND
  {\tt \{tuhin.chakrabarty, smara\}@columbia.edu}\\
  {\tt dghosh@ets.org},
  {\tt npeng@isi.edu}
  }

\date{}

\begin{document}
\maketitle
\begin{abstract}

We propose an unsupervised approach for sarcasm generation based on a non-sarcastic input sentence. Our method employs a retrieve-and-edit framework to instantiate two major characteristics of sarcasm: reversal of valence and semantic incongruity with the context, which could include shared commonsense or world knowledge between the speaker and the listener. While prior works on sarcasm generation predominantly focus on context incongruity, we show that combining valence reversal and semantic incongruity based on commonsense knowledge generates sarcastic messages of higher quality based on several criteria. Human evaluation shows that our system generates sarcasm better than human judges \textit{34\%} of the time, and better than a reinforced hybrid baseline \textit{90\%} of the time.
\end{abstract}

\section{Introduction} \label{section:introduction}

Studies have shown that the use of sarcasm or verbal irony, can  increase creativity on both the speakers and the addressees ~\cite{article}, and can serve different communicative purposes such as evoking humor and diminishing or enhancing  critique \cite{burgers2012verbal}. Thus, developing computational models that generate sarcastic messages could impact many downstream applications, such as better conversational agents and creative or humorous content creation. While most computational work has focused on sarcasm detection \cite{sarc1,sarc6,riloff2013sarcasm,ghosh2015sarcastic,sarc4,muresan2016identification,sarc9,ghoshetal2017role,ghosh2018sarcasm}, research on sarcasm generation is in its infancy \cite{joshi2015sarcasmbot,abhijit}.   
\begin{table}[]
\centering
\small
\begin{tabular}{|@{ }l@{ }|@{ }l@{ }|}
\hline
\textbf{Literal Input 1} & I hate getting sick from fast food.\\
\hline\hline
\textbf{GenSarc1} & I love getting sick from fast food. \\ 
\hline
\begin{tabular}[c]{@{}l@{}} \textbf{GenSarc2}\end{tabular} & \begin{tabular}[c]{@{}l@{}}[I love getting sick from fast food.] [\\Stomach  ache is just an additional side \\effect.]\end{tabular}                     \\ \hline
\textbf{Human 1} & \begin{tabular}[c]{@{}l@{}}Shout out to the Mc donalds for giving\\ me bad food and making me sick right\\ before work in two hours.\end{tabular} \\ \hline
\hline
\textbf{Literal Input 2} & \begin{tabular}[c]{@{}l@{}}I inherited unfavorable genes from my \\mother.\end{tabular}\\
\hline\hline
\textbf{GenSarc3} & \begin{tabular}[c]{@{}l@{}}I inherited great genes from my mother.\end{tabular} \\ 
\hline
\begin{tabular}[c]{@{}l@{}} \textbf{GenSarc4}\end{tabular} & \begin{tabular}[c]{@{}l@{}}[I inherited great genes from my\\ mother.] [Ugly goes down to the bone.]\end{tabular}                     \\ \hline
\textbf{Human 2} & \begin{tabular}[c]{@{}l@{}}Great I inherited all of my mother's \\GOOD genes\end{tabular} \\ \hline
\end{tabular}
\caption{Table showing a literal or non sarcastic input sentence and respective sarcastic outputs. GenSarc1 and GenSarc3 simply reverses the valence, while GenSarc2 and GenSarc4 add commonsense context to create incongruity or enhance the humorous effect.}
\label{table:example1}
\vspace{-1em}
\end{table}
Sarcasm generation is a challenging problem since the generated utterance should have at least five characteristics (a.k.a. ``sarcasm factors'') \cite{burgers2012verbal}: 1)  be evaluative; 2) be based on a reversal of valence between the literal and intended meaning; 3) be based on a semantic incongruity with the context, which can include shared commonsense or world knowledge between the speaker and the addressee; 4)  be aimed at some target, and 5) be relevant to the communicative situation in some way. To simplify the problem, we focus on the task of generating a sarcastic utterance starting from a non-sarcastic utterance that conveys the speaker's intended meaning and that is evaluative. Consider the examples in Table ~\ref{table:example1}. Given the literal input  ``I hate getting sick from fast food"  or ``I inherited unfavorable genes from my mother",  our task is to generate a sarcastic message that would convey this intended literal meaning. In this simplifying task, we are not concerned with the fifth characteristic, while the first and to some degree, the fourth are specified by the input (literal) utterances. 

Given the lack of ``training'' data for the sarcasm generation task, we propose a novel \emph{unsupervised approach} that has three main modules guided by the above mentioned sarcasm factors:
\begin{enumerate}
\item{{\bf Reversal of Valence:} To generate sarcastic utterances that satisfy the second characteristic we identify the evaluative word and use negation or lexical antonyms to generate the sarcastic utterance by reversing the valence (Section \ref{sec:rev1}). For example, given, ``I \textbf{hate} getting sick from fast food'' this module will generate ``I \textbf{love} getting sick from fast food'' (GenSarc1 in Table~\ref{table:example1}).}

\item{ {\bf Retrieval of Commonsense Context:} 
Adding commonsense context could be important to make explicit the semantic incongruity factor (e.g., GenSarc4 vs. GenSarc3 in Table~\ref{table:example1}), or could enhance the humorous effect of the generated sarcastic message (e.g., GenSarc2 vs. GenSarc1 in Table~\ref{table:example1}). 

We propose an approach where retrieved relevant commonsense context sentences are to be added to the generated sarcastic message.  At first, we use a pre-trained language model fine-tuned on the ConceptNet \cite{conceptnet} called COMET \cite{comet} to generate relevant commonsense knowledge. COMET gives us that, ``inherited unfavorable genes from my mother'' causes  \emph{``to be ugly''}  or that ``getting sick from fast food'' causes \emph{``stomach ache''} (Section \ref{section:kb}). The derived commonsense concept is then used to retrieve relevant sentences --- from a corpus ---  that could be added to the sentence obtained through reversal of valence (e.g., ``Stomach ache is just an additional side effect'' in Table~\ref{table:example1}) (Section \ref{section:retrieve}).}

\item{ {\bf Ranking of Semantic Incongruity:} The previous module generates a list of candidate commonsense contexts. Next, we measure \emph{contradiction} between each of these commonsense contexts and the sentence generated by the reversal of valence approach (module 1) and select the commonsense context that received the highest contradiction score. Finally, we concatenate the selected context to the sentence obtained through reversal of valence. Here, conceptually, contradiction detection is aimed to capture the semantic incongruity between the output of valence reversal and its context.  Contradiction scores are obtained from a model trained on the Multi-Genre NLI Corpus \cite{multinli} (Section \ref{sec:ranking}).

}
\end{enumerate}

We test our approach on 150 non-sarcastic utterances randomly sampled from two existing data sets. We conduct human evaluation using several criteria: 1) how \emph{sarcastic} is the generated message; 2) how \emph{humorous} it is; 3) how \emph{creative} it is; and 4) how \emph{grammatical} it is. Evaluation via Amazon's Mechanical Turk (MTurk) shows that our system is better 34\% of the time compared to humans and 90\% of the time compared to a recently published reinforced hybrid baseline \cite{abhijit}. We also present a thorough ablation study of several variations of our system demonstrating that incorporating more sarcasm factors (e.g., reversal of valence, commonsense context, and semantic incongruity) lead to higher quality sarcastic utterances. We make the code and data from our experiments publicly available. \footnote{\url{https://github.com/tuhinjubcse/SarcasmGeneration-ACL2020}}

\section{Related Work} \label{section:related}
\subsection{Sarcasm Generation}
Research on sarcasm generation is in its infancy. \newcite{joshi2015sarcasmbot} proposed \textit{SarcasmBot}, a sarcasm generation system that implements eight rule-based sarcasm generators, each of which generates a certain type of sarcastic expression. 
\newcite{sign} introduced a novel task of sarcasm interpretation, defined as the generation of a non-sarcastic utterance conveying the same message as the original sarcastic one. They use supervised machine translation models for the same in presence of parallel data. However, it is impractical to assume the existence of large corpora for training supervised generative models using deep neural nets; we hence resort to unsupervised approaches.
\newcite{abhijit} employed reinforced neural \emph{seq2seq} learning and information retrieval based approaches to generate sarcasm. Their models are trained using only unlabeled non-sarcastic and sarcastic opinions. They generated sarcasm as a disparity between positive sentiment context and negative situational context. We, in contrast, model sarcasm using semantic incongruity with the context which could include shared commonsense or world knowledge. 

\subsection{Style Transfer}
Prior works looked into \textit{unsupervised} text style/sentiment transfer~\cite{shen2017style,fu2017style,li2018delete}, which transfers a sentence from one style to another without changing the content. This is relevant to the reversal of valence for sarcasm generation. However, these transformations are mainly at the lexical and syntax levels rather than pragmatic level; in contrast, sarcastic utterances often include additional information associated with the context they occur~\cite{regel2009comprehension}, which is beyond text style/sentiment transfer. 

\subsection{Use of Commonsense for Irony Detection}
The study of irony and sarcasm are closely related as sarcasm is defined as, ``the use of verbal irony to mock someone or show contempt''. \newcite{van2018we} addressed the challenge of modeling implicit or prototypical sentiment in the framework of automatic irony detection. They first manually annotated stereotypical ironic situations (e.g., flight delays) and later addressed the implicit sentiment held towards such situations automatically by using both a lexico-semantic commonsense knowledge base and a data-driven method.  They however used it for irony detection, while we are focused on sarcasm generation.\footnote{While we do not directly model the negative intent in sarcasm, the generated output could lead to sarcastic messages rather than just ironic depending on the initial target given in the non-sarcastic message (E.g a sample generation ``Our politicians have everything under control. The nation is in danger of falling into anarchy.")} 


\section{Sarcasm Factors Used in Generation} 
\label{sec:strategies}
A sarcastic utterance must satisfy the sarcasm factors, i.e., the inherent characteristics of sarcasm \cite{attardo2000irony,burgers2012verbal}. In this research, we leverage the use of two particular factors to generate sarcasm. One is the \emph{reversal of valence} and the other is the \emph{semantic incongruity with the context}, which could include shared commonsense or world knowledge between the speaker and the hearer.  
\begin{figure*}[ht]
\centering
\includegraphics[scale=0.30]{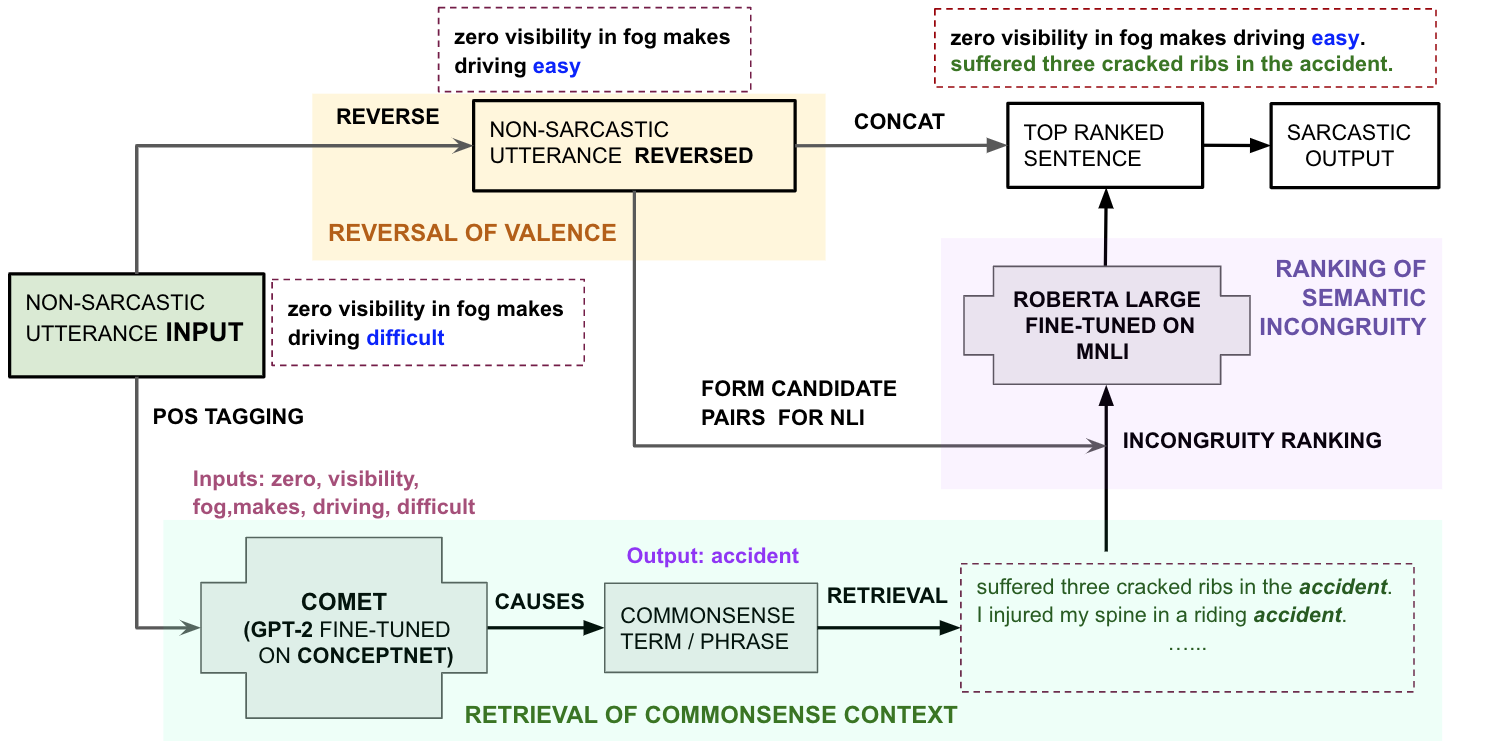}
\caption{\label{figure:pipeline} Our complete pipeline for sarcasm generation. The components with highlighted background denote Reversal of Valence, Retrieval of Commonsense Context and Ranking based on Semantic Incongruity respectively }
\vspace{-1em}
\end{figure*}

\subsection{Reversal of Valence} 
The first key sarcasm factor is the reversal of valence between the literal and the intended meaning \cite{burgers2012verbal}. Reversal of valence can be achieved in two ways: when the literal meaning of the sarcastic message is positive (e.g., ``that is a great outfit'' if the outfit is ugly) or when the literal meaning is negative (e.g., ``that is an ugly dress'' if the dress is really beautiful). Arguably, the former is more likely to appear in sarcastic utterances. As the intended meaning is generally the opposite of its literal meaning in sarcastic utterances \cite{gibbs1986psycholinguistics}, using lexical antonym of negative sentiment words or negation can be used to convert a non-sarcastic utterance to its sarcastic version. For example, given a non-sarcastic utterance ``Zero visibility in fog makes driving \textbf{difficult}", one could identify the evaluative negative word \textit{difficult} and replace it with its antonym \textit{easy}, thereby converting the utterance to the sarcastic ``Zero visibility in fog makes driving \textbf{easy}". Likewise, ``Drunk driving should be taken seriously" can be converted to its sarcastic counterpart, ``Drunk driving should \textbf{not} be taken seriously" by using negation. We propose a generation approach that is able to capture the reversal of valence (Section~\ref{sec:rev1}). 

\subsection{Semantic Incongruity}
 The second sarcasm factor,  semantic incongruity, appears between the literal evaluation and the context, as in the example 
 ``I love getting sick from fast food", where we have semantic incongruity  between the positive word ``love" and the negative situation ``getting sick". However, often, the negative situation is absent from the utterance, and thus additional pragmatic inference is needed to understand the sarcastic intent. For example, the listener might miss the sarcastic intent in ``zero visibility in fog makes driving easy'', where the speaker meant to convey that it can cause \emph{``accidents''}. Adding 
 ``suffered three cracked ribs in an accident.'' makes the sarcastic intent more explicit, while maintaining the acerbic wit of the speaker. 
 In the next section, we propose a novel generation approach that incorporates such relevant commonsense knowledge as context for semantic incongruity (Section~\ref{sec:common} and Section~\ref{sec:ranking}).  



\section{Unsupervised Sarcasm Generation}
\label{sec:approach}

An overview of the sarcasm generation pipeline is shown in Figure \ref{figure:pipeline}. In this section, we detail the three main modules that are designed to instantiate the key sarcasm factors. 

\subsection{Reversal of Valence}
\label{sec:rev1}
As sarcasm is a type of verbal irony used to mock or convey contempt, in most sarcastic messages we encounter a positive sentiment towards a negative situation (i.e., ironic criticism \cite{kreuz2002asymmetries}). This observation is also supported by research on sarcasm detection, particularly on social media. Hence, for our sarcasm generation task, we focus on transforming a literal utterance with negative valence into positive valence.

To implement the reversal of valence, 
as highlighted in the yellow background in Figure \ref{figure:pipeline}, we first identify the evaluative words and replace them with their lexical antonyms using WordNet \cite{miller1995wordnet}. 
As we expect the evaluative words to be negative words, we rely on the word level negative scores obtained from SentiWordNet \cite{esuli2006sentiwordnet}.
In the absence of words with negative polarity, we check if there is the negation word \textit{not} or words ending with \textit{n't} and remove these words. In case there are both negative words and  \textit{not} (or words ending in \textit{n't}), we  handle only one of them. Given the non sarcastic example \textit{``zero visibility in fog makes driving {\bf difficult}''} shown in Figure \ref{figure:pipeline} and which we use as our running example,  the reversal of valence module generates \textit{``zero visibility in fog makes driving {\bf easy}"}. 


\subsection{Retrieval of Commonsense Context}
\label{sec:common}
As discussed before, a straightforward reversal of valence might not generate sarcastic messages that display a clear semantic incongruity, and thus, additional context is needed. We propose an approach to retrieve relevant context for the sarcastic message based on commonsense knowledge. First, we generate commonsense knowledge based on ConcepNet (e.g., ``driving in zero visibility" causes ``accidents'') (Section \ref{section:kb}). Second, we retrieve candidate context sentences that contain the commonsense concept from a retrieval corpus (Section \ref{section:retrieve}) and edit them for grammatical consistency with the input message (Section \ref{section:gr}).

\subsubsection{Commonsense Reasoning} \label{section:kb}
\begin{figure*}[ht]
\centering
\includegraphics[scale=0.18]{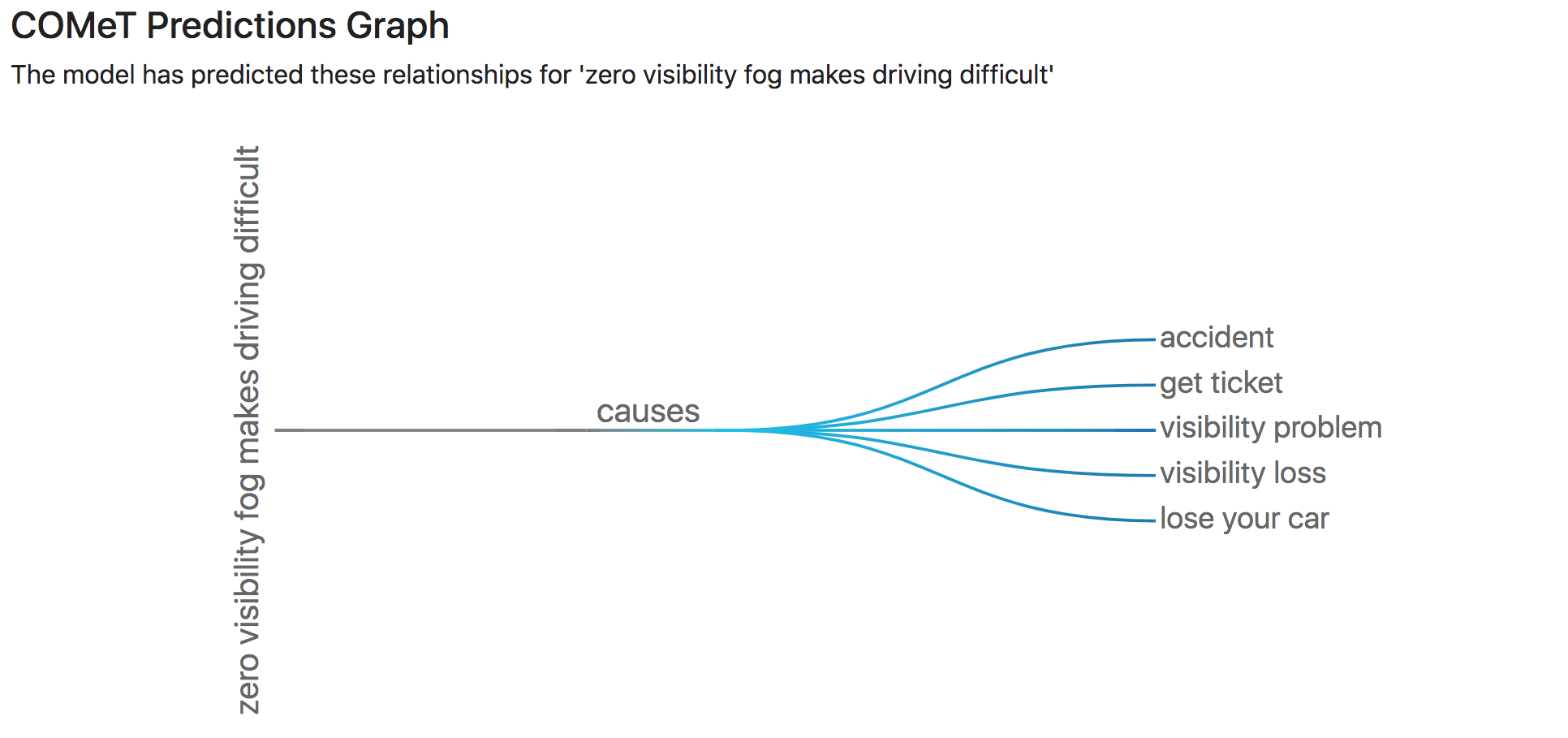}
\caption{\label{figure:comet} Model predictions from COMET. The edges are sorted by probability}
\end{figure*}

We extract nouns, adjectives, adverbs, and verbs from the non-sarcastic input messages and feed them as input to COMET \cite{comet} model to generate commonsense knowledge (highlighted in green background in Figure \ref{figure:pipeline}). COMET is an adaptation framework for constructing commonsense knowledge based on pre-trained language models. It initiates with a pre-trained GPT~\cite{gpt} model and fine-tune on commonsense knowledge tuples (in our case, ConceptNet~\cite{conceptnet}). These tuples provide COMET with the knowledge base structure and relations that must be learned, and COMET adapts the representations that the language model learned from the pre-training stage to add novel nodes to the seed knowledge graph. Our work only leverages the \textbf{causes} relation. 
For instance, from our running example, we first remove the stopwords and then extract nouns, adjectives, adverbs, and verbs including the terms \textit{zero}, \textit{visibility}, \textit{fog},\textit{makes} \textit{driving}, and \textit{difficult} to feed to COMET as inputs. In turn, COMET returns the probable causes with their probability scores. For the running example, COMET returns with the highest probability that these terms may cause an \textbf{accident} (illustrated in Figure \ref{figure:comet}). For further details regarding COMET please see \newcite{comet}. 


\subsubsection{Retrieving Sentences Containing Commonsense Concepts}
\label{section:retrieve}

Once we obtain the most probable output from COMET, the next step is to retrieve sentences containing the commonsense word or phrase from a retrieval corpus. We impose several constraints: (a) the retrieved sentences should contain the commonsense concept at the beginning or at the end; (b) sentence length should be less than twice the number of tokens in the non-sarcastic input to keep a consistency between the length of the non-sarcastic input and its sarcastic version. If none of the commonsense phrase  is  present in the retrieval corpus, we retrieve sentences containing the nouns within the top most phrase. For example, if COMET yields \textit{microwave burger awful} causes the phrase \textbf{food to spoil}, and this phrase does not appear in any sentence in the retrieval corpus, we search for \textit{food} and later replace it in the retrieved sentence with \textit{food to spoil}. COMET often returns output with common phrases such as \textit{you to be}, \textit{you to get}, \textit{person will be}, \textit{you have} which we also removed while keeping the main content word (i.e the commonsense concept)
We use Sentencedict.com, an online sentence dictionary as the retrieval corpus, where one can find high quality sentences for almost every word obeying the above constraints. \footnote{https://sentencedict.com/} 

\subsubsection{Grammatical Consistency}\label{section:gr}


We first check whether the retrieved sentences are consistent with the non-sarcastic input in terms of the pronouns. If the pronouns are mismatched, then we modify the pronoun of the retrieved sentence to match the pronoun of the non-sarcastic input. In case, the non-sarcastic input does not have any pronoun, but the retrieved sentence does, we simply change that pronoun to ``I''. For example, if the non-sarcastic input sentence is \textit{``Ignoring texts is literally the worst part of communication.''} and the retrieved commonsense sentence is \textit{``\textbf{He} has never suffered the torment of rejection.''}, we modify the retrieved sentence to \textit{``\textbf{I} have never suffered the torment of rejection.''} to have consistency among the pronoun use. After correcting the pronouns and proper names (in the same way as pronoun correction), we feed the corrected sentences into the Neural Grammatical Error Corrections System \cite{zhao2019improving} to correct any pronoun or gender specific errors introduced by the replacements.

\subsection{Ranking for Semantic Incongruity}
\label{sec:ranking}
After the grammatical error correction, the next step is to select the best context sentence from the retrieved results. Since we expect the context sentences to be incongruous with the sentence generated by the reversal of valence approach (Section \ref{sec:rev1}), we rank the context sentences by semantic incongruity scores and select the best candidate.

We frame the problem of semantic incongruity based on the Natural Language Inference (NLI)~\cite{nli} task. 
The Multi-Genre NLI \cite{multinli} covers a range of genres of spoken and written text, and supports a distinctive cross-genre generalization, making it an ideal choice as our NLI Dataset. We first fine-tune RoBERTa-large \cite{roberta}, a state-of-the-art pre-trained language model for a 3-way classification (i.e., contradiction, entailment, and neutral) by training on the Multi-NLI dataset. Next, for each retrieved sentence, we treat it as the \emph{premise} and the sentence generated by the reversal of valence as the \emph{hypothesis}, and thus, obtain a contradiction score from the trained model. Finally, the scores obtained for the \textit{contradiction} class are used as a proxy for the degree of \textit{semantic incongruity} and we select the context with the highest score. Figure \ref{figure:pipeline} shows the region with light purple background as our incongruity ranking module. 

\subsection{Implementation Details}

We use the pre-trained COMET model \footnote{https://github.com/atcbosselut/comet-commonsense} for commonsense reasoning with a greedy decoding of five to generate a commonsense phrase and return the topmost that has no lexical overlap with the input. If the generated phrase contains stopwords in the beginning we remove them. For incorporating semantic incongruity, we use the RoBERTa-large model with 355M parameters and fine-tune on MNLI. For grammatical error correction model, we use an open source pre-trained model.\footnote{https://github.com/zhawe01/fairseq-gec}

\section{Experimental Setup} \label{section:exp_setup}
\subsection{Dataset} \label{section:dataset}
\label{sec:data}
\newcite{phrasing} released a dataset of 4,762 pairs of speakers’ sarcastic
messages and hearers’ interpretations by conducting a crowdsourcing experiment. \newcite{sign} introduced a  dataset of 3,000 sarcastic tweets, each interpreted by five human judges and present a novel task of sarcasm interpretation.
Both datasets were collected using the hashtag \textit{\#sarcasm} from Twitter. We merge these two datasets and choose non-sarcastic utterances no longer than 15 words. For each literal non-sarcastic utterance we also keep the corresponding gold sarcastic message, which is useful for evaluation and comparison purposes.
We randomly select 150 utterances as part of the test set (i.e., five times more than the size of the test data in \newcite{abhijit}), while assuring such utterances do not contain high lexical overlap. We allow this constraint to evaluate how our method(s) deal with diverse data.

\begin{table*}[]
\centering
\begin{tabular}{|l|l|l|l|l|}
\hline
System                                              & Sarcasticness & Creativity & Humor & Grammaticality \\ \hline
{State-of-the-art \cite{abhijit}} & 1.63              &  1.60          & 1.50      &   1.46            \\ \hline
Human Generated                                              & \textbf{3.57}              &     3.16       &    \textbf{3.18}   &    3.98            \\ \hline\hline
Reversal of Valence (RV)                           & 3.00              &         2.80   &   2.72    &     \textbf{\color{black}4.29}           \\ \hline
No Reversal of Valence (NoRV)                              & 1.79              & 2.28            &  2.09     &        3.91        \\ \hline
No Semantic Incongruity (NSI)                            &   3.04            &   2.99         & 2.90      &    3.68            \\ \hline
Full Model (FM)                                     &     3.23*          &  \textbf{\color{black}3.24}          &  3.08*     &    3.69            \\ \hline
\end{tabular}
\caption{Average scores for generated sarcasm from all systems as judged by the Turkers. The scale ranges from 1 (\emph{not at all}) to 5 (\emph{very}). For creativity and grammaticality, our models are comparable to human annotation and significantly better than the state-of-the-art ($p<0.001)$. For sarcasticness and humor, the full model is ranked 2nd by a small margin against the human generated message (denoted by *).}
\label{table:example2}
\end{table*}

\subsection{Systems for Experiment} Here, we benchmark the quality of the generated sarcastic messages by comparing multiple systems.
\begin{enumerate}
    \item \textbf{Full Model (FM)}: This model consists of all the three modules aimed at capturing reversal of valence, commonsense context, and semantic incongruity, respectively.
     \item \textbf{Reversal of Valence (RV)}: This model relies only on the reversal of valence component.
     \item \textbf{No Reversal of Valence (NoRV)}: This model only retrieves commonsense context and ranks them based on semantic incongruity. 
     \item \textbf{No Semantic Incongruity (NSI)}: This model relies only on the reversal of valence and  retrieval of commonsense context, without ranking based on semantic incongruity.  A randomly selected retrieved sentence is used.
     \item \textbf{MTS2019}: We make use of the model released by \newcite{abhijit} as it is the state-of-the-art sarcasm generation system.\footnote{https://github.com/TarunTater/sarcasm\_generation}
     \item \textbf{Human (Gold) Sarcasm}: As described in Section~\ref{sec:data}, we have gold sarcasm created by humans for every non-sarcastic utterance.
\end{enumerate}

\label{sec:results}
\subsection{Evaluation Criteria}
BLEU \cite{BLEU} is one of the most widely used automatic evaluation metric for generation tasks such as Machine Translation. However, for creative text generation, it is not ideal to expect significant n-gram overlaps between the machine-generated and the gold-standard utterances. Hence, we performed a human evaluation. We evaluate  a total of 900 generated utterances since our ablation study consisted of six different systems with 150 utterances each.

Sarcasm is  often linked with intelligence, creativity, and wit; thus we propose a set of 4 criteria to evaluate the generated output: (1) \textbf{Creativity} (``How creative are the utterances ?''), (2) \textbf{Sarcasticness} (``How sarcastic are the utterances ?''), (3) \textbf{Humour} (``How funny are the sentences ?'') \cite{skalicky2018linguistic}, and (4) \textbf{Grammaticality} (``How grammatical are the sentences ?''). We design a MTurk task where Turkers were asked to rate outputs from all the six systems. Each Turker was given the non-sarcastic utterance as well as a group of sarcastic utterances generated by all the six systems (randomly shuffled). Each criteria was rated on a scale from 1 (\emph{not at all}) to 5 (\emph{very}). Finally, each utterance was rated by three individual Turkers. 55, 59, 66, and 60 Turkers attempted the HITs (inter-annotator agreement of 0.59, 0.53, 0.47 and 0.66 for the tasks on creativity, sarcasticness, humour and grammaticality, respectively using Spearman's correlation coefficient).

\begin{table}[]
\small
\centering
\begin{tabular}{|l|l|l|l|l|}
\hline
\multirow{2}{*}{Aspect} & \multicolumn{2}{l|}{FM vs Human} & \multicolumn{2}{l|}{FM vs MTS2019} \\ \cline{2-5} 
                        & win\%          & lose\%          & win\%          & lose\%         \\ \hline
Sarcasticness           & 34.0           & \textbf{55.3}            & \textbf{90.0}           & 6.0           \\ \hline
Creativity              &    \textbf{48.0}            &   36.0              &   \textbf{95.3}             &    4.0           \\ \hline
Humor                   &   40.6           &    \textbf{48.0}             &  \textbf{90.0}             &    4.0            \\ \hline
Grammaticality          &  26.6              & \textbf{56.6}                & \textbf{98.0}               &      1.3          \\ \hline
\end{tabular}
\caption{Pairwise comparison between the full model (FM) and
human generated sarcasm, and between the full model (FM) and the state-of-the-art model in \newcite{abhijit}. Win \% (lose \%) is the percentage of the FM gets a higher (lower) average score compared to the other method for the 150 human-rated sentences. The rest are ties.}
\label{table:example3}
\end{table}

\begin{table*}[!ht]
\small
\centering
\begin{tabular}{|l|l|l|l|l|l|l|}
\hline
Non Sarcastic                                                                                             & System & Sarcasm                                                                                                                                                                    & S            & C            & H            & G            \\ \hline
\multirow{6}{*}{\begin{tabular}[c]{@{}l@{}}I inherited \\unfavorable genes \\from my mother.\end{tabular}} & FM     & \begin{tabular}[c]{@{}l@{}}I inherited great genes from my mother. \textbf{\color{black}Ugly} goes  down \\to the bone.\end{tabular}                                                              & \textbf{5.0} & 4.0          & \textbf{3.6} & 3.6          \\ \cline{2-7} 
                                                                                                          & RV  & I inherited great genes from my mother.                                                                                                                                    & 3.0          & 2.6          & 2.0          & 2.3          \\[5pt] \cline{2-7} 
                                                                                                          & NoRV  & \textbf{\color{black}Ugly} goes down to the bone.                                                                                                                                                & 3.0          & 2.6          & 3.0          & \textbf{4.0} \\[5pt] \cline{2-7} 
                                                                                                          & NSI  & \begin{tabular}[c]{@{}l@{}}I inherited great genes from my mother. She makes me \\feel dowdy and \textbf{\color{black}ugly}.\end{tabular}                                                        & 2.6          & 3.6          & 3.0          & \textbf{4.0} \\[5pt] \cline{2-7} 
                                                                                                          & MTS2019   & \begin{tabular}[c]{@{}l@{}}Butch tagging bullies apc seymour good temper\\[5pt] good mentor.\end{tabular}                                                                      & 1.3          & 1.0          & 1.3          & 2.0          \\[5pt] \cline{2-7} 
                                                                                                          & Human  & Great I inherited all of my mother's GOOD genes                                                                                                                            & 2.3          & \textbf{4.3} & 2.0          & 2.6          \\[5pt] \hline
\multirow{6}{*}{\begin{tabular}[c]{@{}l@{}}It is not fun to date \\a  drug addict.\end{tabular}}          & FM     & \begin{tabular}[c]{@{}l@{}}It is fun to date a drug addict. Spent the night in a police \\cell after his \textbf{\color{black}arrest}.\end{tabular}                                             & 4.3          & \textbf{5.0} & \textbf{4.6} & \textbf{5.0} \\[5pt] \cline{2-7} 
                                                                                                          & RV  & It is fun to date a drug addict.                                                                                                                                           & \textbf{5.0} & 2.3          & 2.0          & 4.6          \\[5pt] \cline{2-7} 
                                                                                                          & NoRV  & Spent the night in a police cell after his \textbf{\color{black}arrest}.                                                                                                                         & 1.0          & 1.0          & 2.0          & 2.6          \\[5pt] \cline{2-7} 
                                                                                                          & NSI  & \begin{tabular}[c]{@{}l@{}}It is fun to date a drug addict. The feds completely \\screwed up the \textbf{\color{black}arrest}.\end{tabular}                                                      & 3.3          & 4.3          & 2.0          & 2.6          \\[5pt] \cline{2-7} 
                                                                                                          & MTS2019  & \begin{tabular}[c]{@{}l@{}}Butch is a powerful addict in gente he is \\ an optimist great fun.\end{tabular}                                                                & 2.6          & 2.0          & 1.0          & 1.3          \\[5pt] \cline{2-7} 
                                                                                                          & Human  & Dating a drug addict .. Wouldn't that be fun.                                                                                                                              & 3.0          & 1.6          & 2.6          & 4.0          \\[5pt] \hline
\multirow{6}{*}{\begin{tabular}[c]{@{}l@{}}I hate getting sick \\ from fast food.\end{tabular}}           & FM     & \begin{tabular}[c]{@{}l@{}}I love getting sick from fast food. \textbf{\color{black}Stomach ache}  is just an\\ additional side effect.\end{tabular}                                             & 3.3          & 3.6          & \textbf{5.0} & 3.6          \\[5pt] \cline{2-7} 
                                                                                                          & RV  & I love getting sick from fast food.                                                                                                                                        & 3.3          & 2.6          & 3.6          & \textbf{5.0} \\[5pt] \cline{2-7} 
                                                                                           & NoRV  & \textbf{\color{black}Stomach ache} is just an additional side effect.                                                                                                                            & 1.3          & 2.6          & 3.6          & 3.3          \\[5pt] \cline{2-7} 
                                                                                                          & NSI  & \begin{tabular}[c]{@{}l@{}}I love getting sick from fast food. I ate too  much and got a \\terrible \textbf{\color{black}stomach ache}.\end{tabular}                                              & 2.3          & 3.3          & 4.3          & \textbf{5.0} \\[5pt] \cline{2-7} 
                                                                                                          & MTS2019   & \begin{tabular}[c]{@{}l@{}}I hate love sick to ikes sword lowest **** giving\\ stains giving stains on printers making pound accidents \\work bikinis in\end{tabular} & 1.0          & 1.3          & 1.3          & 1.0          \\[5pt] \cline{2-7} 
                                                                                                          & Human  & \begin{tabular}[c]{@{}l@{}}Shout out to the mcdonalds for giving me  bad food and \\making me sick right before work in two hours.\end{tabular}                         & \textbf{4.0} & \textbf{4.3} & 4.0          & 4.3          \\[5pt] \hline
                                                         
\multirow{6}{*}{\begin{tabular}[c]{@{}l@{}}Burnt popcorn is \\gross.\end{tabular}}           & FM     & \begin{tabular}[c]{@{}l@{}}Burnt popcorn is lovely. The smell made me want to \textbf{\color{black} vomit.}\end{tabular}                                             & \textbf{4.6}          & 3.0          & 3.3 & \textbf{5.0}          \\[5pt] \cline{2-7} 
                                                                                                          & RV  & Burnt popcorn is lovely.                                                                                                                                        & 4.0          & 2.0          & 3.6          & \textbf{5.0} \\[5pt] \cline{2-7} 
                                                                                           & NoRV  & The smell made me want to \textbf{\color{black}vomit.}                                                                                                                            & 1.0          & 2.0          & 3.6          & 4.6         \\[5pt] \cline{2-7} 
                                                                                                          & NSI  & \begin{tabular}[c]{@{}l@{}}Burnt popcorn is lovely. Hold the bag in case I \textbf{\color{black} vomit}.\end{tabular}                                              & 4.3          & 2.3          & 4.3          & \textbf{5.0} \\[5pt] \cline{2-7} 
                                                                                                          & MTS2019   & \begin{tabular}[c]{@{}l@{}} reggae burnt  popcorn lol .\end{tabular} & 2.3          & 1.3          & 2.0          & 1.0          \\[5pt] \cline{2-7} 
                                                                                                          & Human  & \begin{tabular}[c]{@{}l@{}}Gotta love the smell of burnt microwave popcorn.\end{tabular}                         & 3.3 & \textbf{3.3} & \textbf{4.0}          & 4.0          \\[5pt] \hline
\end{tabular}
\caption{Examples of generated outputs from different systems. S, C, H, G represent Sarcasticness, Creativity, Humor and Grammaticality. Text in bolded black represents the commonsense word/phrase obtained from COMET given the non-sarcastic utterance.}
\label{table:analysis}
\end{table*}

\section{Experimental Results}
\subsection{Quantitative Scores}
Table~\ref{table:example2} presents the scores for the above mentioned metrics of different systems averaged over 150 test utterances. 
Our full model as well as the variations that ablated some components improve over the state-of-the-art~\cite{abhijit} on all the criteria. The ablation in Table~\ref{table:example2} shows that our full model is superior to individual modules in terms of sarcasticness, creativity and humor. For grammaticality, we observe that the Turkers scored shorter sentences higher (e.g., RV), which also explains why NoRV model received a higher score than the full model. NoRV otherwise performed worse than all the other variations.

In terms of creativity, our full model attains the highest average scores over all the other models including sarcastic utterances composed by humans. For grammaticality, the reversal of valence model is the best, even better than human generated ones. The performance of the full model is the second best in terms of the sarcasticness and humor, only slightly worse than human-generated sarcasm, showing the effectiveness of our approach that captures various factors of sarcasm. 

\subsection{Pairwise game between Full Model, State-of-the-art and Humans}

Table \ref{table:example3} displays the pairwise comparisons between the full model (FM) and human generated sarcasm, and FM and \newcite{abhijit}, respectively. Given a pair of inputs, we decide win/lose/tie by comparing the average scores (over three Turkers) of both outputs. We see that FM dominates~\newcite{abhijit} on all the metrics and human-generated sarcasm on the creativity metric. For sarcasticness, although humans are better, the FM model still has a 34\% winning rate.

\begin{figure}[t]
\centering
\includegraphics[scale=0.5]{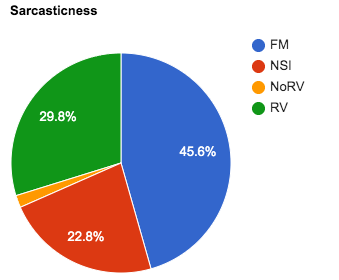}

\caption{\label{figure:pie} Pie chart comparing the success rate of all the variations of our model.}
\end{figure}

\subsection{Ablation Study}
We focus our ablation study on the metric of sarcasticness, as we consider this as the main criterion for the success of generating sarcasm. As shown in Figure~\ref{figure:pie}, our best model (FM) outperforms individual ablation modules. We filtered out 60 examples from the 150 with no ties. The ablation component employing just \textit{Reversal of Valence} is second best for sarcasticness according to Figure ~\ref{figure:pie}.


Further, to understand the extent to which ranking the retrieved sentence based on the degree of incongruity helped generate better sarcasm, we took the outputs from FM and NSI for comparisons. Out of the 150 utterances, 119 times there was no tie. Our best model (FM) wins 66\% of the time while the NSI model wins 34\% of the cases.



\section{Qualitative Analysis} 
Table \ref{table:analysis} demonstrates several generation outputs from different modules associated with human ratings for different criteria. We notice that often one of our modules generate better sarcasm than humans. For instance, for the first and the second example in Table \ref{table:analysis}, all of FM, RV and NSI are better than human generated sarcasm. In general, the generations from the FM model are more humorous, which is also an useful criterion to evaluate sarcasm besides sarcasticness \cite{skalicky2018linguistic}.


We also observe that Turkers consistently rated generations from the FM model more sarcastic than the NSI model suggesting that there is a correlation between human scores of sarcasticness and incongruity. To support this observation, we took the contradiction scores from the RoBERTa model for both best ranked retrieved sentences (FM) and the randomly selected retrieved sentences (NSI). We then computed a correlation between the sarcasticness scores given by the humans and the automatic contradiction scores for both the best ranked retrieved sentences (FM) and the randomly selected retrieved sentences (NSI). For FM model we obtain a higher Pearson correlation coefficient compared to NSI suggesting the important role of incongruency for sarcasm.

\subsection{Limitations}
While our best model combining different sarcasm factors does outperform the system with individual factors, there are sometimes exceptions. We notice, in few cases, the simple reversal of valence (RV) strategy is enough to generate sarcasm. For instance, for the literal input ``It is not fun to date a drug addict", just removing the negation word leads to a full score on sarcasticness without the additional commonsense module. Future work would include building a model that can decide whether just the RV strategy is sufficient or if we need to add additional commonsense context to it.

Although incorporating incongruity ranking is useful, there are several cases when a randomly retrieved message may obtain better sarcasticness score. Table~\ref{table:incongruity} presents such an example. Even though the retrieved message ``Please stop whirling me round; it makes me feel sick." scores lower than ``The very thought of it makes me feel sick.", in terms of incongruity with respect to ``I love being put in the hospital for dehydration", the former received a higher sarcasticness score that suggests the incongruity scores obtained from NLI are not perfect.

The ordering of the commonsense context and the valence reversed sentence is predetermined in our generation. Specifically, we always append the retrieved commonsense context after the valence reversed output. Changing the order can sometimes make the sarcasm better and more humorous. The reason for our current ordering choice is that we always treat the valence reversed version as \textit{hypothesis} and the commonsense retrieved sentence as \textit{premise} for the NLI model. We attempted reversing the order in preliminary experiments but received poor scores from the entailment model. In future, we would like to generate more diverse sarcasm that are not tied to a fixed pattern.

Finally, the generations are dependent on COMET and thus the quality will be governed by the accuracy of the COMET model. 

\begin{table}[t]
\small
\centering
\begin{tabular}{|p{0.4cm}|l|}
 \hline
NSI    & \begin{tabular}[c]{@{}l@{}}I love being put in the hospital for dehydration. \\ Please stop whirling me round; it makes me\\  feel \textbf{sick}.\end{tabular} \\ \hline
FM   & \begin{tabular}[c]{@{}l@{}}I love being put in the hospital for dehydration. \\ The very thought of it makes me feel \textbf{sick}.\end{tabular}               \\ \hline
\end{tabular}
\caption{Sarcastic Generation from (FM) and (NSI) where NSI scores higher for sacrasticness}
\label{table:incongruity}
\vspace{-1.5em}
\end{table}

\section{Conclusion}
We address the problem of unsupervised sarcasm generation that models several sarcasm factors including reversal of valence and semantic incongruity with the context. 
The key contribution of our approach is the modeling of commonsense knowledge in a retrieve-and-edit generation framework. 
A human-based evaluation based on four criteria shows that our generation approach significantly outperforms a state-of-the-art model. Compared with human generated sarcasm, our model shows promise particularly for creativity, humor and sarcasticness, but less for grammaticality. 
A bigger challenge in sarcasm generation and more generally, creative text generation, is to capture the difference between creativity (novel but well-formed material) and nonsense (ill-formed material). Language models conflate the two, so developing methods that are nuanced enough to recognize this difference is key to future progress.

\section*{Acknowledgments}

This work was supported in part by the MCS program under Cooperative Agreement N66001-19-2-4032 and the CwC program under Contract W911NF-15-1-0543 with the US Defense Advanced Research Projects Agency (DARPA). The views expressed are those of the authors and do not reflect the official policy or position of the Department of Defense or the U.S. Government.
The authors would like to thank Christopher Hidey, John Kropf, Anusha Bala and Christopher Robert Kedzie for useful discussions. The authors also thank members of PLUSLab at the University Of Southern California and the anonymous reviewers for helpful comments.

\bibliography{anthology,acl2020}
\bibliographystyle{acl_natbib}

\end{document}